\documentclass[sigconf]{acmart}
\pdfoutput=1
\usepackage{subfigure}
\usepackage{algorithmic}

\usepackage[linesnumbered,ruled, vlined]{algorithm2e}
\usepackage{multirow}
\usepackage[misc]{ifsym}

\AtBeginDocument{%
  \providecommand\BibTeX{{%
    \normalfont B\kern-0.5em{\scshape i\kern-0.25em b}\kern-0.8em\TeX}}}

\setcopyright{acmcopyright}
\copyrightyear{2021}
\acmYear{2021}
\acmDOI{10.1145/1122445.1122456}

\acmConference[WSDM '21]{ International Conference on Web Search and Data Mining}{March 08--12, 2021}{Jerusalem, Israel}
\acmBooktitle{WSDM '21:{International Conference on Web Search and Data Mining,
  March 08--12, 2021, Jerusalem, Israel}
\acmPrice{15.00}
\acmISBN{978-1-4503-XXXX-X/18/06}
}

\newcommand{\stitle}[1]{\vspace{0.75ex}\noindent{\bf #1}}

\begin{document}

\title{Learning to Drop: Robust Graph Neural Network via Topological Denoising}
\author{Dongsheng Luo$^{1*}$, Wei Cheng$^{2*}$, Wenchao Yu$^2$, Bo Zong$^2$, Jingchao Ni$^2$,}
\author{Haifeng Chen$^{2}$, Xiang Zhang$^1$}
\affiliation{%
  \institution{$^1$Pennsylvania State University, $^2$NEC Labs America}
}
\email{
{dul262,xzz89}@psu.edu, {weicheng,wyu,bzong,jni,haifeng}@nec-labs.com
}
\thanks{*Equal Contribution}

\begin{abstract}
Graph Neural Networks (GNNs) have shown to be powerful tools for graph analytics. The key idea is to recursively propagate and aggregate information along edges of the given graph. Despite their success, however, the existing GNNs are usually sensitive to the quality of the input graph. Real-world graphs are often noisy and  contain task-irrelevant edges, which may lead to suboptimal generalization performance in the learned GNN models. In this paper, we propose PTDNet, a parameterized topological denoising network,  to improve the robustness and generalization performance of GNNs by learning to drop task-irrelevant edges. PTDNet prunes task-irrelevant edges by penalizing the number of edges in the sparsified graph with parameterized networks. To take into consideration of  the topology of the entire graph, the nuclear norm regularization is applied to impose the low-rank constraint on the resulting sparsified graph for better generalization. PTDNet can be used as a key component in GNN models to improve their performances on various  tasks, such as node classification and link prediction. Experimental studies on both synthetic and benchmark datasets show that PTDNet can improve the performance of GNNs significantly and the performance gain becomes larger for more noisy datasets.

\end{abstract}

\maketitle

\vspace{-0.1cm}
\section{Introduction}
\label{sec:intro}

In recent years, we have witnessed a dramatic increase in our ability to extract and collect data from the physical world. In many applications, data with complex structures are connected for their interactions and are naturally represented as graphs~\cite{wang2017uncovering, ni2018co, van2013wu}. 
Graphs are powerful data representations but are challenging to work with because they require modeling both node feature information as well as rich relational information among nodes~\cite{sen2008collective, carlson2010toward, yang2016revisiting, hamilton2017inductive}. To tackle this challenge, various Graph Neural Networks (GNNs) have been proposed to aggregate information from both graph topology and node features~\cite{hamilton2017inductive, kipf2016semi,velivckovic2017graph, xu2018powerful,zhang2020adaptive}. GNNs model node features as messages and propagate them along the edges of the input graph. During the process, GNNs compute the representation vector of a node by recursively aggregating and transforming representation vectors of its neighboring nodes. 
Such methods have achieved state-of-the-art performances in various tasks, including node classification and link prediction~\cite{zhang2018deep,zhou2018graph}.

Despite their success, GNNs are vulnerable to the quality of the given graph due to its recursively aggregating schema. It is natural to ask: \textit{is it necessary to aggregate all neighboring nodes? If not, is there a principled way to select which neighboring nodes are not needed to be included?} In many real-world applications, graph data exhibit complex topology patterns. Recent works~\cite{rong2020dropedge,zhao2019pairnorm} have shown that GNNs are greatly \textit{over-smoothed} as edges can be pruned without loss of accuracy. Besides, GNNs are easily aggregating task-irrelevant information, leading to \textit{over-fitting} which weakens the generalization ability. Specifically,
from the local perspective, a node might be linked to nodes with task-specific ``noisy'' edges. Aggregating information from these nodes would impair the quality of the node embedding and lead to unwanted prediction in the downstream task. From the global view, nodes located at the boundary of clusters are connected to nodes from multiple communities. Overwhelming information collected from their neighbors would dilute the true underlying patterns. 

\begin{figure}[t]
    \centering
    \includegraphics[width=3.0in]{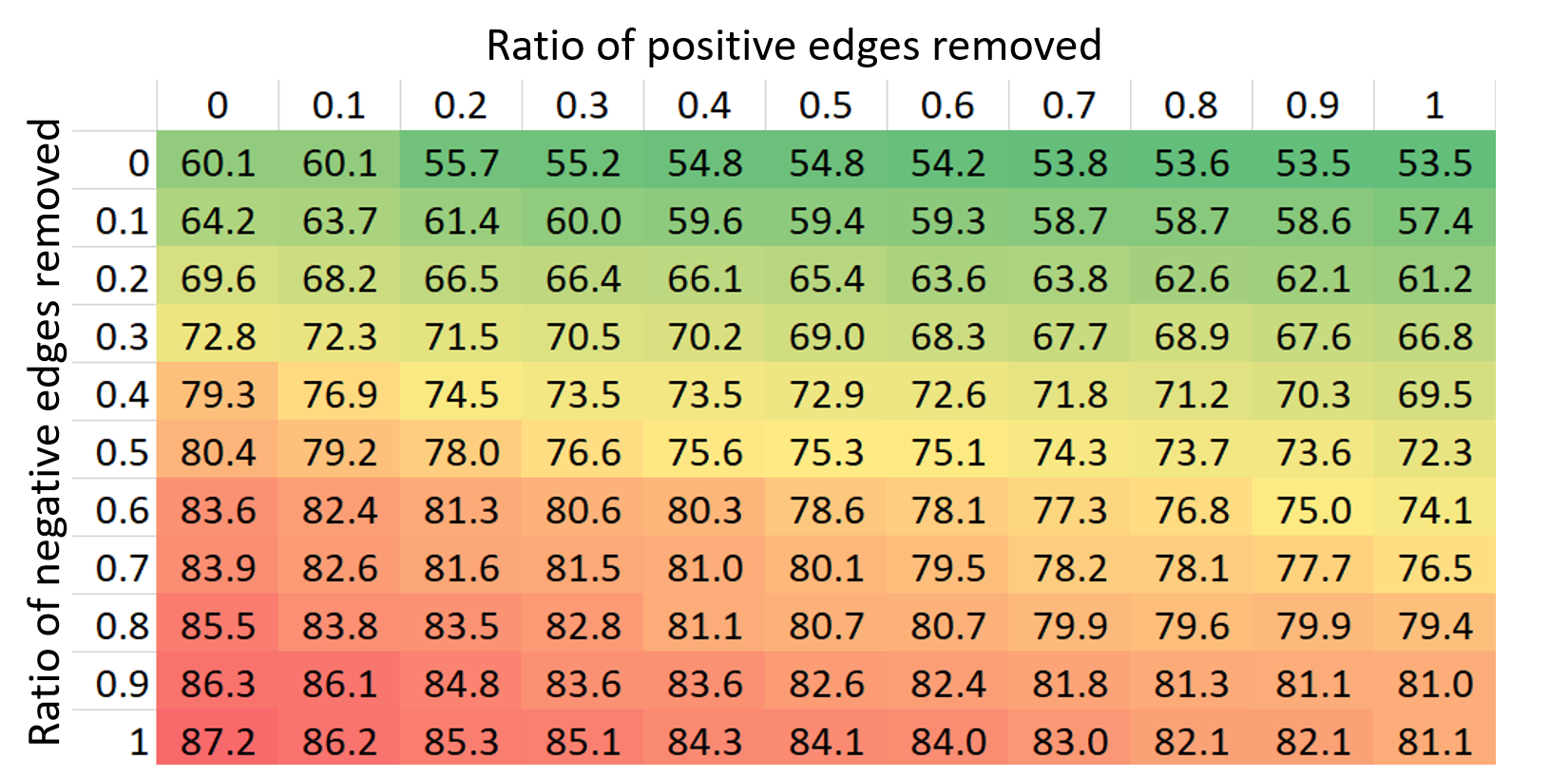}
    \caption{Performance of GCN w.r.t. removing positive and negative edges on Cora.}
    \label{fig:pos_neg_gcn}
    \vspace{-0.6cm}
\end{figure}

As a motivating example, we consider a benchmark dataset Cora~\cite{sen2008collective}. We denote the edges connecting nodes with the same label as positive edges, otherwise, as negative edges. 
Table~\ref{tab:stat1} shows the statistics of different edges. 
It is reasonable to consider that passing messages through positive edges leads to high quality node representation, while information aggregated along negative edges impair the performance of GNNs~\cite{Hou2020Measuring}.  We adopt Graph Convolutional Network(GCN)~\cite{kipf2016semi}, a representative GNN, as an example to verify this intuition. We randomly delete some positive and negative edges and conduct GCN on the resulting graphs. 
As shown in Fig.~\ref{fig:pos_neg_gcn}, the performance of GCN increases with more negative edges removed. 
\begin{table}[t]
    \centering
    \caption{Statistics of positive and negative edges on Cora.}
    \label{tab:stat1}
    \begin{small}
    \begin{tabular}{c|c|c|c|c}
    \hline
        Dataset  & \# Nodes   & \# Edges  &  \# Pos. Edges & \# Neg. Edges\\
    \hline
        Cora     & 2,708      & 5,429     & 4,418          &1,011\\ 
    \hline
    \end{tabular}
    \end{small}
    \vspace{-0.6cm}
\end{table}


Topological denoising is a promising solution to address the above-mentioned challenge by removing ``noisy'' edges~\cite{Entezari2020all,wang2019learning}. 
By denoising the input graph, we can prune away task-irrelevant edges to avoid aggregating unnecessary information in GNNs. Besides, it can also help improve the robustness and alleviate the over-smoothing problem inherent of GNNs~\cite{rong2020dropedge}. The idea of topological denoising is not new. In
fact, this line of thinking has motivated GAT~\cite{velivckovic2017graph} that aggregates neighboring nodes with weights from attention mechanism and thus to some extent alleviate the problem. Other existing methods aim to extract smaller subgraphs from the given graphs to preserve pre-defined properties or randomly remove/sample edges during the training process to prevent GNNs from over-smoothing~\cite{shin2019sweg,rong2020dropedge, hamilton2017inductive,wang2020inductive}.
However, within unsupervised settings, subgraphs sampled from these approaches may be suboptimal for downstream tasks and also lack persuasive rationales to explain the outcomes of the model for the task. Instead, the task-irrelevant ``noisy'' edges should be specific to the downstream objective. Besides, in real-life graphs, node contents and graph topology provide complementary information to each other. Denoising process should take both information into consideration, which is overlooked by existing methods.

In this paper, we propose a Parameterized Topological Denoising network (PTDNet) to enhance the performance of GNNs. We use deep neural networks, considering both structural and content information as inputs, to learn to drop task-irrelevant edges in a data-driven way. PTDNet prunes the graph edges by penalizing the number of edges in the sparsified graph with parameterized networks~\cite{louizos2017learning, xue2020learning}. The denoised graphs are then fed into GNNs for robust learning. The introduced sparsity in the neighboring nodes aggregation has a variety of merits: 1) a sparse aggregation is less complicated and hence generalizes well~\cite{journalsGirosi98}; 2) it can facilitate interpretability and help infer task-relevant neighbors. Considering the combinatorial nature of the denoising process, we relax the discrete constraint with continuous distributions that could be optimized efficiently with backpropagation, enabling PTDNet to be compatible with various GNNs in both transductive and inductive settings, including GCN~\cite{kipf2016semi}, Graph Attention Network~\cite{velivckovic2017graph}, GraphSage~\cite{hamilton2017inductive}, etc. In PTDNet, the denoising networks and GNN are jointly optimized in an end-to-end fashion. Different from conventional methods that remove edges randomly or based on pre-defined rules, the denoising process of PTDNet is guided by the supervision of the downstream objective in the training phase.

To further concern the global topology, the nuclear norm regularization is applied to impose low-rank constraint on the resulting sparsified graph for better generalization. Due to the discontinuous nature of the rank minimization problem, PTDNet smooths the constraint with the nuclear norm, which is the tightest convex envelope of the rank~\cite{recht2010guaranteed}. This regularization denoises the input graph from the global topology perspective by removing edges connecting multiple communities to improve generalization ability and robustness of GNNs~\cite{Entezari2020all}. Experimental results on both synthetic and benchmark datasets demonstrate that PTDNet can effectively enhance the performance and robustness of GNNs.

\section{Related Work}
\label{sec:related}
\textbf{GNNs} are powerful tools to investigate the graph data with node contents. GNN models utilize the message passing mechanism to encode both graph structural information and node features into vector representations. These vectors are then used to node-level or graph-level downstream tasks. GNNs were initially proposed in~\cite{gori2005new}, and extended in~\cite{scarselli2008graph}. These methods learn node representations by iteratively aggregating neighbor information until reaching a static state. 
Inspired by the success of convolutional neural networks (CNNs) in computer vision, graph convolutional networks in the graph spectral domain were proposed based upon graph Fourier transform~\cite{bruna2013spectral}. Multiple extensions were further proposed~\cite{kipf2016semi,defferrard2016convolutional,scarselli2008graph,li2018adaptive,velivckovic2017graph,ma2019disentangled}.
The express power of GNNs were analyzed in~\cite{xu2018powerful}. DropEdge and PairNorm investigated the over-smoothing problem of stacking multiple GNN layers~\cite{rong2020dropedge, zhao2019pairnorm}.

\textbf{Graph Sparsification and Sampling.} Conventional graph sparsification approximates the large input graph with a sparse subgraph to enable efficient computation, and at the same time preserve certain properties. Different notions have been extensively studied including pairwise distances betweenness~\cite{chew1989there}, sizes of all cuts~\cite{benczur1996approximating}, node degree distributions~\cite{eden2018provable}, and spectral properties~\cite{hermsdorff2019unifying,arora2019differentially}. These methods remove edges only based upon the structural information, which limits their power when combining with GNNs. Besides, without supervised feedback from the downstream task, these approaches may generate subgraphs with suboptimal structural properties. NeuralSparse~\cite{zheng2020robust} learns $k$-neighbor subgraphs for robust graph representation learning by selecting at most $k$ edges for each nodes. The $k$-neighbor assumption however limits its learning power and may lead to suboptimal performance  in generalization.

Recently, graph sampling has been investigated in GNNs for fast computation and better generalization capacity, including neighbor-level~\cite{hamilton2017inductive}, node-level~\cite{chen2018fastgcn, huang2018adaptive,graphsaint}, and edge-level sampling methods~\cite{rong2020dropedge}. Unlike these methods that randomly sample edges in the training phase, PTDNet utilizes parametrized networks to actively remove task-specific noisy edges. With supervised guidance from downstream objective, the generated subgraphs benefit GNNs in not only robustness but also accuracy and interpretability.  Besides, PTDNet has better generalization capacity as the parametrized networks can be used for inductive inference.


\vspace{-0.2cm}
\section{Notations and preliminaries}
\label{sec:notations}
\textbf{Notations.}  In general, we use lowercase, bold uppercase, and bold lowercase letters for scalars, matrices, and vectors, respectively. For example, we use $\mathbf{Z}$ to denote a matrix, whose $i,j$-th entry is denoted by  $z_{ij}$, a lowercase character with subscripts. Let $G=(\mathcal{V},\mathcal{E})$ represent the input graph with $n$ nodes, where $\mathcal{V},\mathcal{E}$ stand for its node/edge set, respectively. The adjacency matrix of $G$ is denoted by $\mathbf{A}\in \mathbb{R}^{n\times n}$. Node features are denoted by matrix $\mathbf{X} \in \mathbb{R}^{n\times m}$ with $m$ as the dimensionality of node features. 
We use $\mathbf{Y}$ to denote the labels in the downstream task. For instance, in the node classification task, $\mathbf{Y}\in \mathbb{R}^{n\times c}$ represents node labels, where $c$ is the number of classes. 

\textbf{GNN layer.} Applying a GNN layer consists of the propagation step and the output step~\cite{zhou2018graph}. At the propagation step, the aggregator first computes the message for each edge. For an edge $(v_i,v_j)$, the aggregator takes the representations of $v_i$ and $v_j$ in previous layer as inputs, denoted by $\mathbf{h}_i^{t-1}$, and $\mathbf{h}_j^{t-1}$, respectively. Then, the aggregator collects messages from local neighborhoods for each node $v_i$. At the output step, the updater computes its new hidden representation, denoted by  $\mathbf{h}_v^t$. 

GNN models adopt message passing mechanisms to propagate and aggregate information along the input graph to learn node representations. The performances can be heavily affected by the quality of the input graph. Messages aggregated along ``noisy'' edges may decrease the quality of node embeddings. Overwhelming information from multiple communities put GNNs at the risk of over-smoothing, especially when multiple GNN layers are stacked~\cite{zhao2019pairnorm,rong2020dropedge,loukas2020what}. Existing methods either utilize graph sparsification strategies to extract subgraphs or randomly sample graphs to enhance the robustness of GNNs. Basically, they are conducted in an unsupervised way, limiting their ability to filter out \textit{task-specific} noisy edges. 

The core idea of PTDNet is to actively filter out task-specific noisy edges in the input graph with a parameterized network. It consists of the denoising network and general GNNs. GNNs can be applied under both inductive and transductive settings. 
We first give an overview of PTDNet in Sec.~\ref{sec:framework}, followed by details of the denoising network in Sec.~\ref{sec:denoising}. To enhance the generalization ability of PTDNet, we further introduce the low-rank constraint on resulting graphs and provide smoothing relaxation to achieve an end-to-end model.

\section{The PTDNet}
\label{sec:local}
\subsection{The overall architecture}
\label{sec:framework}

The architecture of PTDNet is shown in Fig.~\ref{fig:framework}. It consists of two major components, the denoising networks and the GNNs. The denoising network is a multi-layer network that samples a subgraph from a learned distribution of edges. PTDNet is compatible with most existing GNNs, such as GCN~\cite{kipf2016semi}, GraphSage~\cite{hamilton2017inductive}, GAT~\cite{velivckovic2017graph}, GIN~\cite{xu2018powerful}, etc.
With relaxations, the denoising network is differentiable and can be jointly optimized with GNNs guided by supervised downstream signals. 
\subsection{The denoising network}
\label{sec:denoising}
\subsubsection{Graph edge sparsification}
The goal of the denoising network is to generate a subgraph filtering out task-irrelevant edges for GNN layers. For the $l$-th GNN layer, we introduce a binary matrix $\mathbf{Z}^l \in \{0,1\}^{|\mathcal{V}|\times |\mathcal{V}|}$, with $z^l_{u,v}$ denoting whether the edge between node $u$ and $v$ is present (0 indicates noisy edge).

Formally, the adjacency matrix of the resulting subgraph is $ \mathbf{A}^l = \mathbf{A}\odot \mathbf{Z}^l$
,where $\odot$ is the element-wise product.
One way to reduce noisy edges with the least assumptions about $\mathbf{A}^l$ is to directly penalize the number of non-zero entries in $\mathbf{Z}^l$ of different layers.
\begin{equation}
\setlength{\abovedisplayskip}{2pt}
\setlength{\belowdisplayskip}{2pt}
\label{eq:l0}
\begin{small}
    \sum_{l=1}^L ||\mathbf{Z}^l||_0=\sum_{l=1}^L\sum_{(u,v)\in \mathcal{E}} \mathbb{I}[z^l_{u,v}\neq 0],
\end{small}
\end{equation}
where $\mathbb{I}[\cdot]$ is an indicator function, with $\mathbb{I}[True]=1$ and  $\mathbb{I}[False]=0$, $||\cdot||_0$ is the $\ell_0$ norm. There are $2^{|\mathcal{E}|}$ possible states of $\mathbf{Z}^l$. Because of its nondifferentiability and combinatorial nature, optimizing this penalty is computationally intractable. 
Therefore, we consider each binary number $z^l_{u,v}$ to be drawn from a Bernoulli distribution parameterized by $\pi^l_{u,v}$, i.e., $z^l_{u,v} \sim Bern(\pi^l_{u,v})$. The matrix of $\pi_{u,v}^l$'s is denoted by $\Pi^l$. Then, penalizing the non-zero entries in $\mathbf{Z}^l$, i.e., the number of edges being used, can be reformulated as regularizing $\sum_{(u,v)\in \mathcal{E}} \pi^l_{u,v}$
~\cite{louizos2017learning}.


Since $\pi_{u,v}^l$ is optimized jointly with the downstream task, it describes the task-specific quality of the edge $(u,v)$. A small value of $\pi_{u,v}^l$ indicates the edge $(u,v)$ is more likely to be noise and should be with small weight or even be removed in the following GNN. Although the regularization of the reformulated form is continuous, the adjacency matrix of the resulting graph is still generated by a binary matrix $\mathbf{Z}^l$. The expected cost of downstream task could be modeled as $L(\{\Pi^l\}^L_{l=1}) = \mathbb{E}_{\mathbf{Z}^1\sim p(\Pi^1),\cdots,\mathbf{Z}^L\sim p(\Pi^L)}f(\{\mathbf{Z}^l\}^L_{l=1},\mathbf{X})$. To minimize the expected cost via gradient descent, we need to estimate the gradient $\nabla_{\Pi^l}\mathbb{E}_{\mathbf{Z}^1\sim p(\Pi^1),\cdots,\mathbf{Z}^L\sim p(\Pi^L)}f(\{\mathbf{Z}^l\}^L_{l=1},\mathbf{X})$, $l \in [1,2,\cdots,L]$. Existing methods adopt various estimators to approximate the gradient, including score function~\cite{williams1992simple}, straight-through~\cite{bengio2013estimating}, etc. However, these methods suffer from either high variance or biased gradients~\cite{mnih2014neural}. In addition, to make PTDNet suitable for the inductive setting and enhance generalization ability, a parameterized method for modeling $\mathbf{Z}^l$ should be adopted.

\vspace{-0.25cm}
\subsubsection{Continuous relaxation with parameterized networks}
To efficiently optimize subgraphs with gradient methods, we adopt the reparameterization trick~\cite{jang2016categorical} and relax the binary entries $z_{u,v}^l$ from being drawn from a Bernoulli distribution to a deterministic function $g$ of parameters $\alpha_{u,v}^l \in \mathbb{R}$ and an independent random variable $\epsilon^l$. That is  $z_{u,v}^l = g(\alpha_{u,v}^l,\epsilon^l)$.
\begin{equation}
\setlength{\abovedisplayskip}{3pt}
\setlength{\belowdisplayskip}{3pt}
    \begin{small}
    \nabla_{\alpha_{u,v}^l}\mathbb{E}_{\epsilon^1,...,\epsilon^L} f(\{g,\mathbf{X}\})
    = \nabla_{\alpha_{u,v}^l}\mathbb{E}_{\epsilon^1,...,\epsilon^L} \left[ \frac{\partial f}{\partial g} \frac{\partial g}{\partial \alpha_{u,v}^l}\right].
\end{small}
\end{equation}

\begin{figure}[tb]
\vspace{-0.2cm}
    \centering
    \subfigure[Framework of PTDNet.]{\includegraphics[width=2.3in]{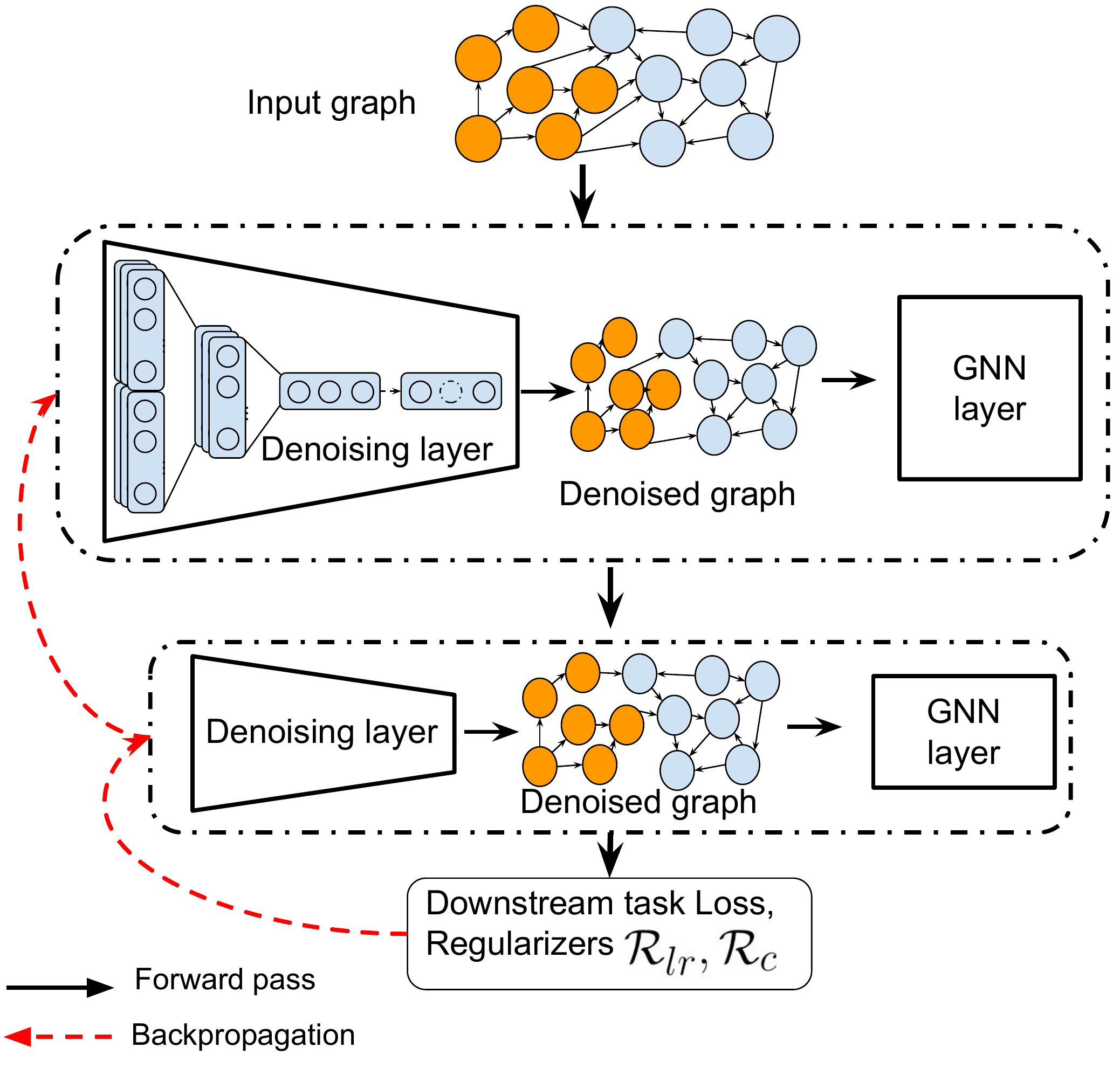}\label{fig:framework}}\hspace{0.2in} 
    \subfigure[Computation for low-rank constraint.]{\includegraphics[width=2.3in]{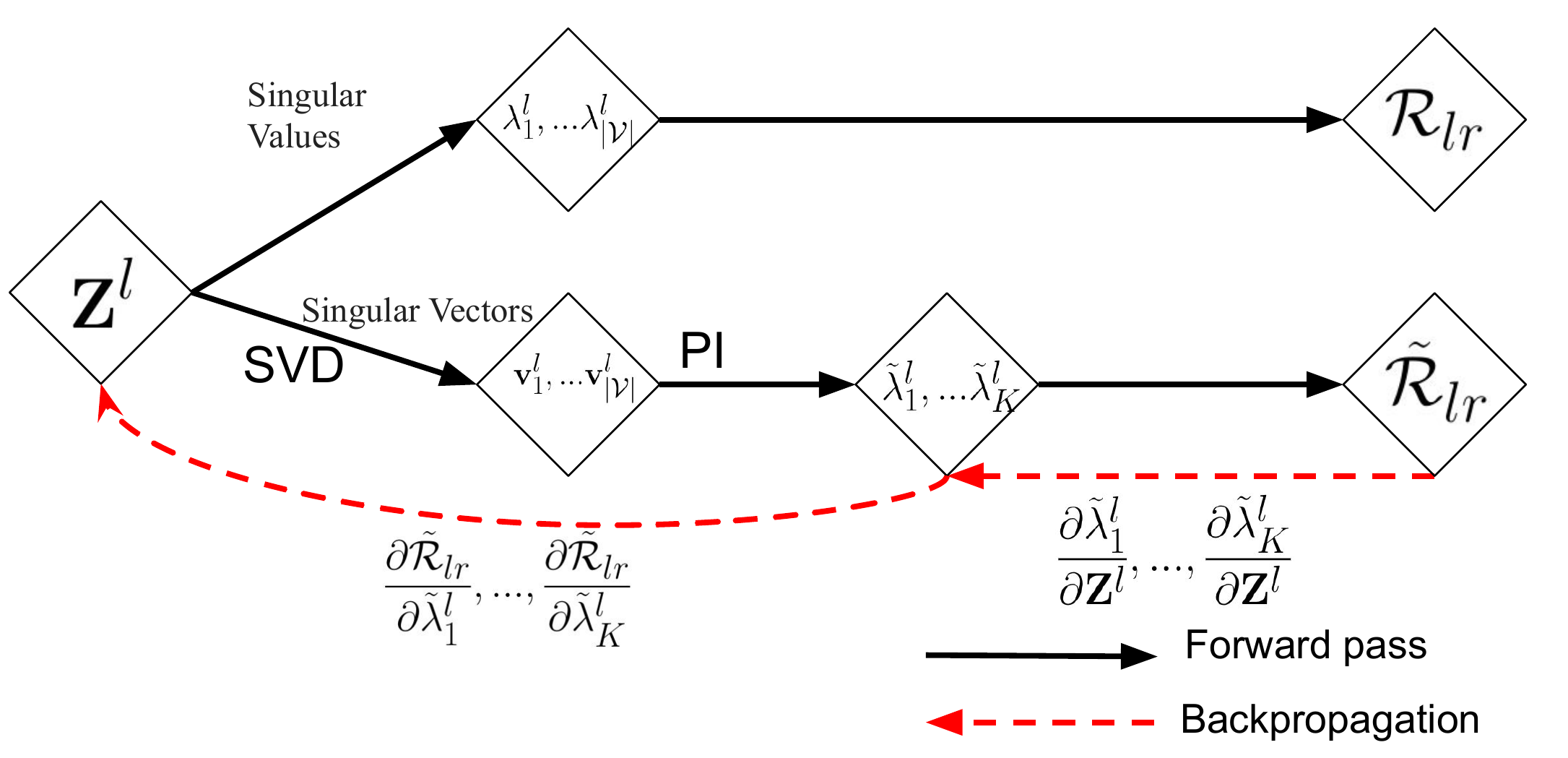}\label{fig:lowrank}}
    \vspace{-0.35cm}
    \caption{The PTDNet model.}
    \label{fig:method}
    \vspace{-0.5cm}
\end{figure}

To enable the inductive setting, we should not only figure out which edges but also why they should be filtered out. To learn to drop, for each edge $(u,v)$, we adopt parameterized networks to model the relationship between the task-specific quality $\pi_{u,v}^l$ and the node information including node contents and topological structure. In the training phase, we jointly optimize denoising networks and GNNs. In the testing phase, the input graphs could also be denoised with the learned denoising networks. Since we need to compute a subgraph of the input graph, the time complexity of the denoising network in the inference phase is linear to the number of edges $O(|\mathcal{E}|)$.

Following ~\cite{velivckovic2017graph}, we adopt deep neural networks to learn the
parameter $\alpha_{u,v}^l$ that controls whether to remove the edge ($u,v$). Without loss of generality, we focus on a node $u$ in the training graph. Let $\mathcal{N}_u$ be its neighbors. For the $l$-th GNN layer, we calculate $\alpha_{u,v}^l$ for node $u$ and $v\in\mathcal{N}_u$ with $\alpha^l_{uv}=f^l_{\theta^l}(\mathbf{h}^l_u,\mathbf{h}^l_v)$,
where $f_{\theta^l}^l$ is an MLP parameterized by $\theta^l$. 
To get $z^l_{u,v}$, we utilize the concrete distribution along with hard sigmoid function~\cite{maddison2016concrete,louizos2017learning}.
First, we draw  $s_{u,v}^l$ from a binary concrete distribution with $\alpha^l_{uv}$ parameterizing the location~\cite{maddison2016concrete,jang2016categorical}. Formally,
\begin{equation}
\begin{small}
\setlength{\abovedisplayskip}{3pt}
\setlength{\belowdisplayskip}{3pt}
    \epsilon \sim \text{Uniform}(0,1), \: \quad
  s_{u,v}^l = \sigma ((\log \epsilon-\log(1-\epsilon)+\alpha^l_{uv})/\tau),
\end{small}  
\end{equation}
where $\tau \in \mathcal{R}^+$ indicates the temperature and $\sigma(x)=\frac{1}{1+e^{-x}}$ is the sigmoid function. With $\tau>0$, the function is smoothed with a well-defined gradient $\frac{\partial s_{u,v}^l}{\partial \alpha^l_{uv}}$, enabling efficient optimization of the parameterized denoising network.

Since the binary concrete distribution has a range of (0,1), to encourage the weights for task-specific noisy edges to be exactly zeros, we first extend the range to $(\gamma,\zeta)$, with $\gamma<0$ and $\zeta>1$~\cite{louizos2017learning}. Then, we compute $z^l_{u,v}$ by clipping the negative values to 0 and values larger than 1 to 1.
\begin{equation}
\begin{small}
\setlength{\abovedisplayskip}{3pt}
\setlength{\belowdisplayskip}{3pt}
    \label{eq:extend}
    \bar{s}_{u,v}^l = t(s_{u,v}^l) = s_{u,v}^l(\zeta-\gamma)+\gamma, \quad
    z^l_{u,v} = \text{min}(1,\text{max}(\bar{s}_{u,v}^l,0)).
\end{small}
\end{equation}

Within the above formulation, the constraint on the number of non-zero entries in $\mathbf{Z}^l$ in Eq.(~\ref{eq:l0}) can be reformulated with 
\begin{equation}
\begin{small}
\setlength{\abovedisplayskip}{3pt}
\setlength{\belowdisplayskip}{3pt}
       \mathcal{R}_c = \sum_{l=1}^L \sum_{(u,v)\in \mathcal{E}} (1-\mathbb{P}_{\bar{s}^l_{u,v}}(0|\theta^l)).
\end{small}
\end{equation}
$\mathbb{P}_{\bar{s}^l_{u,v}}(0|\theta^l)$ is the cumulative distribution function (CDF) of $\bar{s}^l_{u,v}$.

As shown in~\cite{maddison2016concrete}, the density of $s^l_{u,v}$ is 
\begin{equation}
    p_{s^l_{u,v}}({x}) = \frac{\tau \alpha_{u,v}^l x^{-\tau-1}(1-x)^{-\tau-1}}{(\alpha_{u,v}^l x^{-\tau}+(1-x)^{-\tau})^2}.
\end{equation}
The CDF of variable $s^l_{u,v}$ is 
\begin{equation}
     \mathbb{P}_{s^l_{u,v}}(x) = \sigma((\log x - \log(1-x))\tau-\alpha_{u,v}^l).
\end{equation}
Since the function $\bar{s}_{u,v}^l = t(s_{u,v}^l)$ in Eq.~(\ref{eq:extend}) is monotonic. The probability density function of $\bar{s}_{u,v}^l$ is 
\begin{equation}
    \begin{aligned}
    &p_{\bar{s}^l_{u,v}}(x) =  p_{s^l_{u,v}}(t^{-1}(x))|\frac{\partial}{\partial x } t^{-1}(x)|\\
    &= \frac{(\zeta-\gamma)\tau \alpha_{u,v}^l (x-\gamma)^{-\tau-1}(\zeta-x)^{-\tau-1}}{(\alpha_{u,v}^l(x-\gamma)^{-\tau}+(\zeta-x)^{-\tau})^2}.
   \end{aligned}
\end{equation}
Similarly, we have the CDF of $\bar{s}^l_{u,v}$
\begin{equation}
\begin{aligned}
    &\mathbb{P}_{\bar{s}^l_{u,v}}(x) = \mathbb{P}_{s^l_{u,v}}(t^{-1}(x))\\
    &=\sigma((\log(x-\gamma)-\log(\zeta-x))\tau-\alpha_{u,v}^l).
\end{aligned}
\end{equation}
By setting $x=0$, we have the 
\begin{equation}
    \mathbb{P}_{\bar{s}^l_{u,v}}(0|\theta^l)=\sigma(\tau\log\frac{-\gamma}{\zeta}-\alpha_{u,v}^l).
\end{equation}
Algorithm~\ref{alg:training} summarizes the overall training of PTDNet.

\vspace{-0.2cm}
\begin{algorithm}[htbp]
\small
    \caption{PTDNet algorithm}\label{alg:training}
    \begin{algorithmic}[1]
        \STATE {\bfseries Input:} Training Graph $G=(\mathcal{V},\mathcal{E})$, node features $\mathbf{X}$, number of GNN layers $L$, labels $\mathbf{Y}$ for downstream task.
        \STATE {\bfseries Output:} Predicted labels for downstream task
        \FOR{each minibatch}
        \FOR{$l \leftarrow 1$ {\bfseries to} $L$}
            \STATE $G_d=(\mathcal{V},\mathcal{E}_d) \leftarrow$ subgraph of $G$ sampled by $l$-th denoising network.
            \STATE Feed $G_d$ into the following GNN layer.
            \STATE $\mathbf{H}^l \leftarrow$ hidden representations updated by GNN layer.
        \ENDFOR
         \STATE $\{\hat{\mathbf{y}}_v|v\in\mathcal{V} \}\leftarrow $  prediction with $\mathbf{H}^L$.
         \STATE Compute $loss(\hat{\mathbf{Y}},\mathbf{Y})$ and regularizors.
        \STATE  Update parameters of GNN and denoising networks.
        \ENDFOR
    \end{algorithmic}
\end{algorithm}
\vspace{-0.1cm}
\begin{algorithm}[htbp]
\small
    \centering
    \caption{Forward pass to compute nuclear norm}
     \label{alg:lowrank}
    \begin{algorithmic}[1]
        \STATE {\bfseries Input:} Adjacency $\mathbf{A}^l$, approximate hyper-parameter $K$
        \STATE {\bfseries Output:} nuclear norm loss $\tilde{\mathcal{R}}_{lr}$
        \FOR{$l \leftarrow 1$ {\bfseries to} $L$}
        \STATE $(\mathbf{U}^l)\Lambda^l(\mathbf{V}^l)^* \leftarrow \text{SVD}(\mathbf{A}^l)$ \hfill \textcolor{blue}{$\triangleright$ dismiss gradients at backpro}.
        \STATE $\mathbf{B} \leftarrow (\mathbf{A}^l)^*\mathbf{A}^l$
        \FOR{$i \leftarrow 1$ {\bfseries to} $K$}
            \STATE $\mathbf{v}^l_i \leftarrow $PI($\mathbf{B},\mathbf{v}_i^l $)
            \STATE $\tilde{\lambda}^l_i \leftarrow  \sqrt{[(\mathbf{v}^l_i)^T\mathbf{B}\mathbf{v}^l_i]/[(\mathbf{v}^l_i)^T\mathbf{v}^l_i]}$
            \STATE $\mathbf{B} \leftarrow \mathbf{B}-\mathbf{B}\mathbf{v}^l_i(\mathbf{v}^l_i)^T$ \hfill \textcolor{blue}{$\triangleright$ deflation}
        \ENDFOR
    \ENDFOR
    \STATE $\tilde{\mathcal{R}}_{lr} \leftarrow \sum_{l=1}^L \sum_{i=1}^{K}|\tilde{\lambda}^l_i|$
    \end{algorithmic}
\end{algorithm}

\normalsize

\subsection{The low-rank constraint}
\label{sec:lowrank}
In the previous section, we introduced parameterized networks to remove the task-specific noisy edges from the local neighborhood perspective. In real-life graph data, nodes from multiple classes can be divided into different clusters. Intuitively, nodes from different topological communities are more likely with different labels~\cite{journalscorr}. Hence, edges connecting multiple communities are highly possible noise for GNNs. Based upon this intuition, we further introduce a low-rank constraint on the adjacency matrix of the resulting subgraph to enhance the generalization capacity and robustness, since the rank of the adjacency matrix reflects the number of clusters. This regularization denoises the input graph from the global topology perspective by encouraging the denoising networks to remove edges connecting multiple communities such that the resulting subgraphs to have dense connections within communities while sparse between them~\cite{kanada2018low}. Recent work also shows that graphs with low rank are more robust to network attacks~\cite{Entezari2020all,jin2020graph}.

Formally, the straightforward regularizer $\mathcal{R}_{lr}$ for low-rank constraint of PTDNet is $\sum_{l=1}^L \text{Rank}(\mathbf{A}^{l})$, where $\mathbf{A}^l$ is the adjacency matrix for the $l$-th GNN layer. It has been shown in previous studies that the matrix rank minimization problem (RMP) is NP-hard~\cite{david1995algorithms}. We approximately relax the intractable problem with the nuclear norm, which is the convex surrogate for RMP problem~\cite{friedland2018nuclear}. The nuclear norm of a matrix is defined as the sum of its singular
values. It is a convex function that can be optimized efficiently. Besides, previous studies have shown that in practice, nuclear norm constraints produce very low-rank solutions~\cite{Entezari2020all,friedland2018nuclear, recht2010guaranteed}. With nuclear norm minimization, the regularizer is
\begin{equation}
\begin{small}
\setlength{\abovedisplayskip}{3pt}
\setlength{\belowdisplayskip}{3pt}
    \mathcal{R}_{lr} = \sum_{l=1}^L||\mathbf{A}^l||_* = \sum_{l=1}^L \sum_{i=1}^{|\mathcal{V}|}|\lambda^l_i|,
\end{small}
\end{equation}
where $\lambda_i^l$ is the $i$-th largest singular values of graph adjacency matrix $\mathbf{A}^l$. 

Singular value decomposition (SVD) is required to optimize the nuclear norm regularization. However, SVD may lead to unstable results during backpropagation. As shown in~\cite{ionescu2015matrix}, the partial derivatives of the nuclear norm requires computing of a matrix $\mathbf{M}^l$ with elements 
\begin{equation}
\begin{small}
\setlength{\abovedisplayskip}{3pt}
\setlength{\belowdisplayskip}{2pt}
    m^l_{ij} = \left\{
    \begin{array}{ll}
         1/((\lambda^l_i)^2-(\lambda^l_j)^2),& i\neq j \\
         0,& i=j 
    \end{array}.
    \right.
\end{small}
\end{equation}

When $(\lambda^l_i)^2-(\lambda^l_j)^2$ is small, the partial derivatives become very large, leading to an arithmetic overflow. Besides, the gradient-based optimization on SVD is time-consuming.
The Power Iteration (PI) method with deflation procedure is one way to solve this problem~\cite{nakatsukasa2013stable,ortega1990numerical,wang2019backpropagation}. PI approximately computes the largest eigenvalue and the dominant eigenvector of the matrix $(\mathbf{A}^l)^*\mathbf{A}^l$with an iterative procedure
from a randomly initiated vector. $(\mathbf{A}^l)^*$ stands for the transpose-conjugate matrix of $\mathbf{A}^l$. The largest singular value of $\mathbf{A}^l$ is then the square root of the largest eigenvalues.
To calculate other eigenvectors, the deflation procedure is involved to iteratively remove the projection of the input matrix on this vector. However, PI may output inaccurate approximations if two eigenvalues are close to each other. The situation becomes worse when we consider eigenvalues near zero. Besides, with randomly initiated vectors, PI may need more iterations to get a precise approximation.

To address the problem, we combine SVD and PI~\cite{wang2019backpropagation} and further relax the nuclear norm to Ky Fan $K$-norm~\cite{fan1951maximum}, which is the sum of top $K$, $1\leq K \ll |\mathcal{V}|$, largest singular values. Fig. ~\ref{fig:lowrank} shows the forward pass and backpropagation of the nuclear norm. In the forward pass, as shown in Algorithm~\ref{alg:lowrank}, SVD is used to calculate singular values, left and right singular vectors. Then we get the nuclear norm as the regularization loss. In order to minimize the nuclear norm, we utilize the power iteration to compute top $K$ singular values, denoted by $\tilde{\lambda}^l_1,...\tilde{\lambda}^l_{K}$. Note that the PI process does not update the values in singular vectors and singular values. It only serves to compute the gradients during backpropagation, which is shown with red dot lines in Fig.~\ref{fig:lowrank}.
We estimate the nuclear norm with $\tilde{\mathcal{R}}_{lr} = \sum_{l=1}^L \sum_{i=1}^{K}|\tilde{\lambda}^l_i|$. $\tilde{\mathcal{R}}_{lr}$ is a lower bound function of $\mathcal{R}_{lr}$ with gap
\begin{equation}
\setlength{\abovedisplayskip}{3pt}
\setlength{\belowdisplayskip}{3pt}
\begin{small}
    \begin{aligned}
    \label{eq:topk}
    &\mathcal{R}_{lr}- \tilde{\mathcal{R}}_{lr} = \sum_{l=1}^L \sum_{i=K+1}^{|\mathcal{V}|}|\tilde{\lambda}^l_i| = \sum_{l=1}^L \sum_{i=K+1}^{|\mathcal{V}|}|\lambda^l_i|\leq (|\mathcal{V}|-K)\sum_{l=1}^L |\lambda^l_{K+1}|.
    \end{aligned}
\end{small}
\end{equation}
It is obvious that $\lceil\frac{|\mathcal{V}|}{K}\rceil\tilde{\mathcal{R}}_{lr}$ is the upper bound of $\mathcal{R}_{lr}$. We dismiss the constant coefficient and minimize $\tilde{\mathcal{R}}_{lr}$ as the low-rank constraint.

\section{Experimental study}
\label{sec:exp}
In this section, we empirically evaluate the robustness and effectiveness of PTDNet with both synthetic and benchmark datasets. First, we apply PTDNet to popular GNN models for node classification on benchmark datasets. Second, we evaluate the robustness of PTDNet by injecting additional noise. Moreover, we also provide insight into the denoising process by checking the edges removed by PTDNet. We also conduct comprehensive experiments to uncover insights of PTDNet, including empirically demonstrating the effects of regularizers, parameter study, analyzing the over-smoothing problem inherent in GNNs, and applying PTDNet to another downstream task, i.e., link prediction.

\begin{table}[h]
    \centering
    \vspace{-0.1cm}
    \caption{Dataset statistics}
    \vspace{-0.1cm}
    \label{tab:stat}
    \begin{small}
    \begin{tabular}{c|c|c|c|c}
    \hline
        Dataset  &Cora      &Citeseer   & Pubmed  & PPI \\
    \hline
        Nodes    &2,708     &3,327      & 19,717    &56,944     \\
        Edge     &5,429     &4,732      & 44,338    &818,716     \\
        Fea. &1,433     &3,703      & 500           &50     \\
        Classes  &7         &6          &3               &121\\
Train.   &140       &120        &100          &44,906\\
Val. &500       &500        &500         &6,514\\
Test.    &1,000     &1,000      &1,000     &5,524\\
\hline
    \end{tabular}
    \vspace{-0.1cm}
    \end{small}
\end{table}

\vspace{-0.2cm}
\subsection{Experimental setup}
\vspace{-0.1cm}
\stitle{Datasets.} 
Four benchmark datasets are adopted in our experiments. Cora, Citeseer, and Pubmed are citation graphs where each node is a document and edges describe the citation relationship. A document is assigned with a unique label based on its topic. Node features are bag-of-words representations of the documents. We follow the standard train/val/test splits in~\cite{kipf2016semi,velivckovic2017graph} with very scarce labelled nodes, which are different from the full-supervised setting in DropEdge~\cite{rong2020dropedge}. PPI contains graphs describing protein-protein interaction in different human tissues. Positional gene sets, motif gene sets, and immunological signatures are used as node features. Gene ontology sets are used as labels. The statistics of these datasets are listed in Table~\ref{tab:stat}.

\stitle{Implementations \& metrics.}
We consider three representative GNNs as backbones, including GCN~\cite{kipf2016semi}, GraphSage~\cite{hamilton2017inductive}, GAT~\cite{velivckovic2017graph}. Note that our model is a general framework that is compatible with diverse GNN models. Recent sophisticated models can also be combined with our framework to improve their performances and robustness. Achieving SOTA performances by using a complex architecture is not the main research point of this paper. 
We compare with most recent state-of-the-art sampling and sparsification methods, DropEdge~\cite{rong2020dropedge} and NeuralSparse~\cite{zheng2020robust}. For GraphSage, we use the mean aggregation. We follow the experimental setting in~\cite{rong2020dropedge} to 
perform a random hyper-parameter search for each model. For each setting, we run 10 times and report the average results. Parameters are tuned via cross-validation. 
We also include a variant of PTDNet by removing the low-rank constraint as the ablation study.
For single-label classification datasets, including Cora, Citeseer, and Pubmed, we evaluate the performance with accuracy~\cite{kipf2016semi}. For PPI, we evaluate with micro-F1 scores~\cite{hamilton2017inductive}.

All experiments are conducted on a Linux machine with 8 NVIDIA Tesla V100 GPUs, each with 32GB memory. CUDA version is 9.0 and Driver Version is 384.183.  All methods are implemented with Tensorflow 1.12.

\subsection{Effectiveness evaluation}
\vspace{-0.1cm}
\begin{table}[htbp]
 \vspace{-0.2cm}
\centering
\begin{scriptsize}
    \centering
        \caption{Node classification.}
    \label{tab:results}
    \vspace{-0.2cm}
    \begin{tabular}{c|c|c|c|c|c}
    \hline
        Backbone &Method   &Cora              & Citeseer          & Pubmed             & PPI \\
    \hline
         \multirow{5}{*}{GCN}&Basic                  &0.811 $\pm$ 0.015  & 0.703 $\pm$ 0.012             & 0.790 $\pm$ 0.020              &0.660 $\pm$ 0.024   \\ 
         &DropEdge       &0.809 $\pm$ 0.035          & 0.722 $\pm$ 0.032             & 0.785 $\pm$ 0.043              &0.606 $\pm$ 0.041   \\
         &NeuralSparse   &0.821 $\pm$ 0.014          & 0.715 $\pm$ 0.014            & 0.788 $\pm$ 0.018              &0.651 $\pm$ 0.014   \\
         &PTDNet-wl      &0.824 $\pm$ 0.018          & 0.717 $\pm$ 0.170             & 0.791 $\pm$ 0.012             &0.752 $\pm$ 0.017    \\
         &PTDNet         &0.828 $\pm$ 0.026          & 0.727 $\pm$ 0.018             & \textbf{0.798} $\pm$ 0.024              &0.803 $\pm$ 0.008   \\
         \hline
         \multirow{5}{*}{GraghSage} &Basic           &0.792 $\pm$  0.027        & 0.676 $\pm$ 0.023             & 0.767 $\pm$ 0.020             &0.618 $\pm$ 0.014     \\
         &DropEdge  &0.787 $\pm$ 0.023         &0.670 $\pm$ 0.031              & 0.748 $\pm$ 0.026            &0.610 $\pm$ 0.035     \\
         &NeuralSparse  &0.793 $\pm$ 0.021          &  0.674 $\pm$ 0.011             & 0.751 $\pm$ 0.021              &0.626 $\pm$ 0.023   \\
         &PTDNet-wl &0.794 $\pm$ 0.026         &0.678 $\pm$ 0.022              & 0.770 $\pm$ 0.024             &0.645  $\pm$ 0.020   \\
         &PTDNet &0.803    $\pm$  0.019        &0.679 $\pm$ 0.018              & 0.771 $\pm$ 0.010          &0.648 $\pm$ 0.025    \\
         \hline 
         \multirow{5}{*}{GAT} &Basic &0.830 $\pm$ 0.007   & 0.721 $\pm$ 0.009      & 0.790 $\pm$ 0.008                    &0.973 $\pm$ 0.012      \\
         &DropEdge       &0.832 $\pm$ 0.040     &0.709 $\pm$ 0.020      &0.779 $\pm$  0.019    &0.850 $\pm$ 0.038      \\
         &NeuralSparse   &0.834 $\pm$ 0.015    & 0.724 $\pm$ 0.026     & 0.780 $\pm$ 0.017   &0.921 $\pm$ 0.018   \\
         &PTDNet-wl     &0.837 $\pm$ 0.022      & 0.723 $\pm$ 0.014     &0.792  $\pm$ 0.014   &0.978 $\pm$ 0.018      \\
         &PTDNet        &\textbf{0.844} $\pm$   0.023   & \textbf{0.737} $\pm$ 0.031         &0.793 $\pm$ 0.015                &\textbf{0.980} $\pm$ 0.022       \\
\hline
    \end{tabular}
     \vspace{-0.25cm}
\end{scriptsize}
\end{table}
Table~\ref{tab:results} summarizes the results on different datasets. PTDNet-wl is the variant of PTDNet by removing the low-rank constraint for ablation study. The comparison results demonstrate that by including the denoising part, PTDNet achieves the state-of-the-art or matched performance across different benchmarks.  Specifically, 1) comparing to basic GNNs, PTDNet-wl can improve the performance and generalization capacity by including denoising process to GNNs. 2) In PTDNet, we further include the low-rank constraint to denoise the input graph from the global perspective. As discussed in Sec.~\ref{sec:lowrank}, graphs with low ranks are more robust to complex structures. It encourages to remove the edges across different clusters and helps to alleviate the over-smoothing problem. 3) PTDNet utilizes a parameterized method to actively remove task-irrelevant edges or decrease their weights. The denoising networks can also be used in the testing phase, which shows a better generalization capacity. While DropEdge only works in the training phase to randomly remove edges. These explain the reason why PTDNet outperforms DropEdge. 4) PTDNet outperforms recent work -- NeuralSparse because NeuralSparse constrains the extracted subgraphs to be $k$-neighbor graphs. The $k$-neighbor assumption however limits its learning power and may lead to suboptimal performance in generalization. Moreover, NeuralSparse does not consider the low-rank constraints on the resulting sparsified graph, thus achieves worse generalization performance.

\begin{figure*}[h]
    \centering
    \subfigure[GCN]{\includegraphics[width=1.5in]{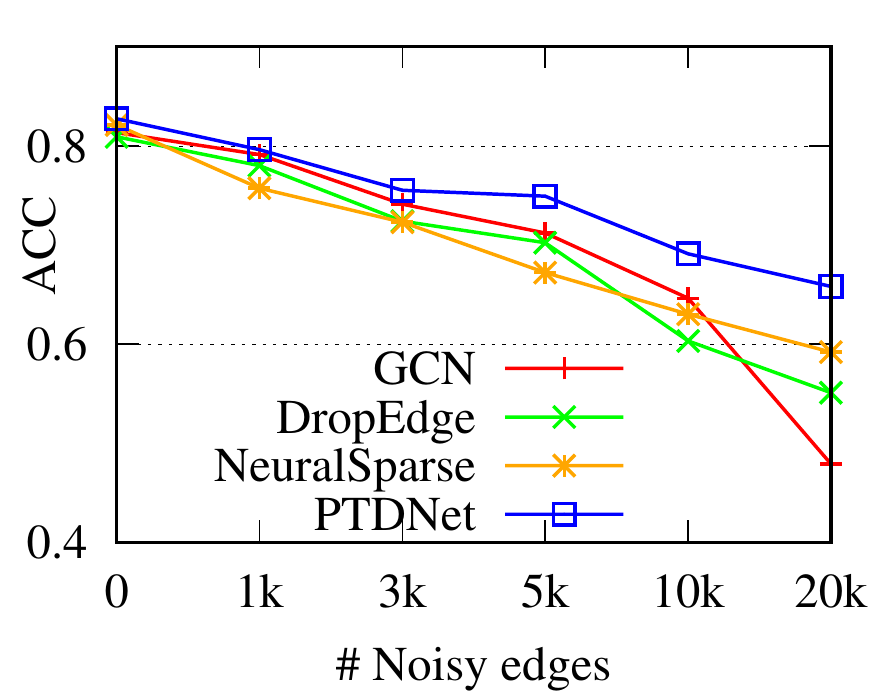}\label{fig:exp2:gcn}}\hspace{0.6cm}
    \subfigure[GraphSage]{\includegraphics[width=1.5in]{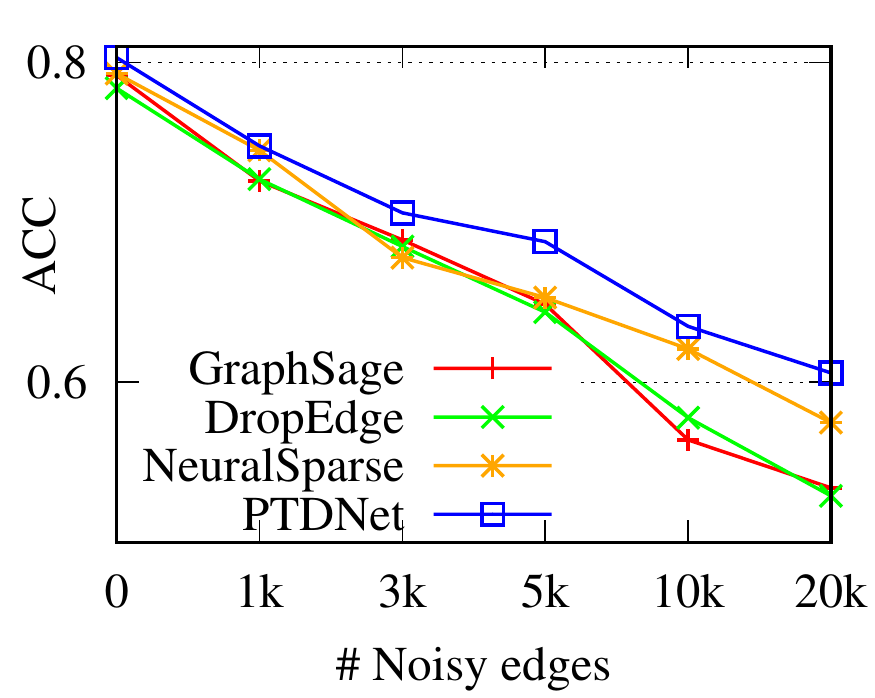}\label{fig:exp2:graphsage}}\hspace{0.6cm}
    \subfigure[{GAT}]{\includegraphics[width=1.5in]{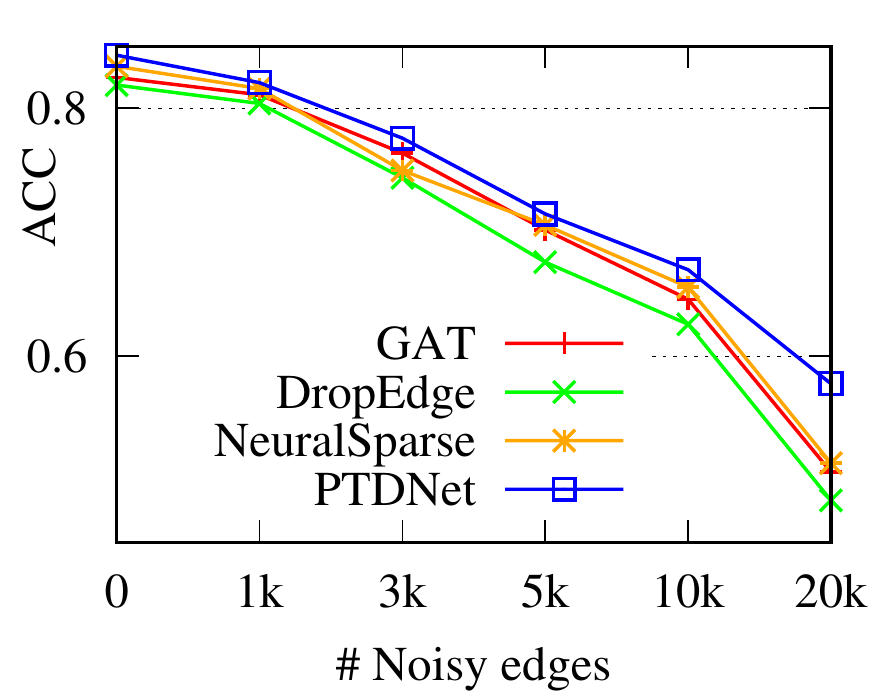}\label{fig:exp2:gat}}
    \vspace{-0.4cm}
    \caption{Accuracy performance with noisy edges in Cora.}
    \label{fig:cora_gcn_noisy}
    \vspace{-0.5cm}
\end{figure*}

\vspace{-0.1cm}
\subsection{Robustness evaluation}
In this part, we evaluate the robustness of PTDNet by manually including noisy edges. We use the Cora dataset and randomly connect $N$ pairs of previously unlinked nodes with $N$ ranging from 1000 to 20,000. We compare the proposed method to baselines with all there backbones. 
Performances are shown in Fig.~\ref{fig:cora_gcn_noisy}. We have the following observations. 1) PTDNet consistently outperforms DropEdge, NeuralSparse and the basic backbones with various numbers of noisy edges. The comparison demonstrates the robustness of PTDNet. 2) DropEdge randomly samples a subgraph for GNN layers. In most cases, DropEdge reports worse performances than the original backbones. The comparison demonstrates that a random sampling strategy used in DropEdge is vulnerable to noisy edges. 3) 
NeuralSparse selects noise edges guided by task signals thus achieves better results than basic backbones. However, it selects at most $k$ edges for each nodes, which may lead to suboptimal performance. 4) The margins between results of PTDNet and basic backbones become large when more noise are injected. Specifically, PTDNet relatively improves the accuracy scores by 37.37\% for GCN, 13.4\% for GraphSage, and 16.1\% for GAT with 20,000 noisy edges.

\vspace{-0.1cm}
\subsection{On denoising process}
\label{sec:exp:syn}
In this section, we use controllable synthetic datasets to analyze the denoising process of PTDNet, which has 5 labels and 30 features per node. We first randomly sample five 30-dimensional vectors as the centroids, one for each label. Then, for each label, we sample nodes from a Gaussian distribution with the centroid as the mean. The variance, which controls the quality of content information is set to 80. The number of nodes for each label is drawn from another Gaussian distribution $\mathcal{N}(200,25)$.

To validate GNNs on fusing node content and topology, we build a graph containing \textit{complementary} information with node features. Specifically, we use the distance between a node and the centroid node of its label as the metric to evaluate the quality of the node feature.  The probability that it connects to another node with a different label is positively proportional to the feature quality. The resulting graph contains 1,018 nodes and 4,945 edges. 
We randomly select 60/20/20\% nodes for training/validation/testing, respectively.

\begin{figure}[h]
\vspace{-0.47cm}
    \centering
    \subfigure[Training edges]{\includegraphics[width=1.4in]{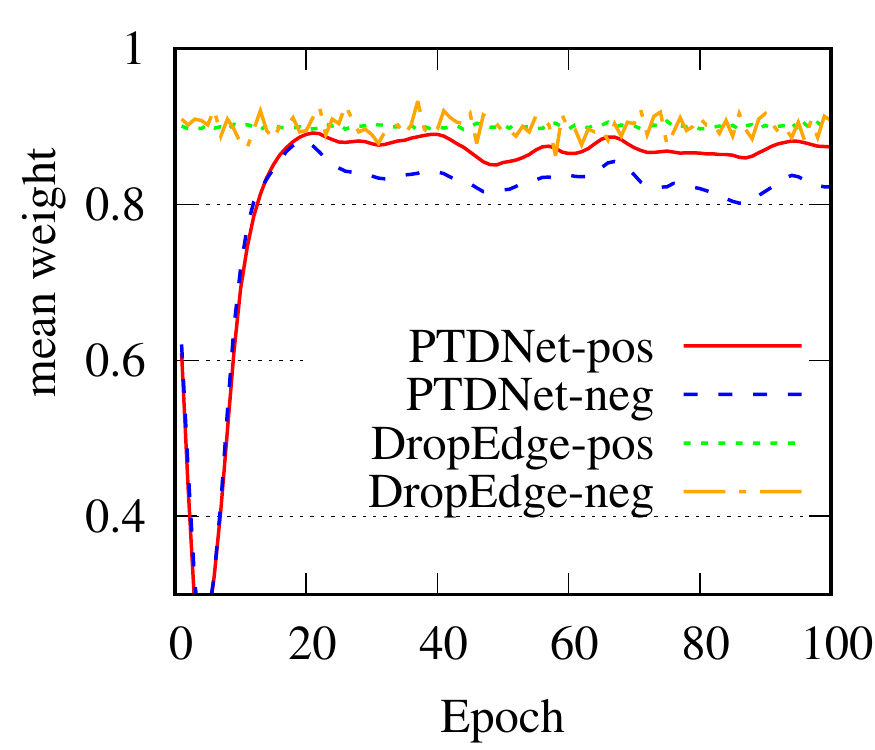}\label{fig:exp3:train}}\hspace{-0.05in}
    \subfigure[{Testing edges}]{\includegraphics[width=1.4in]{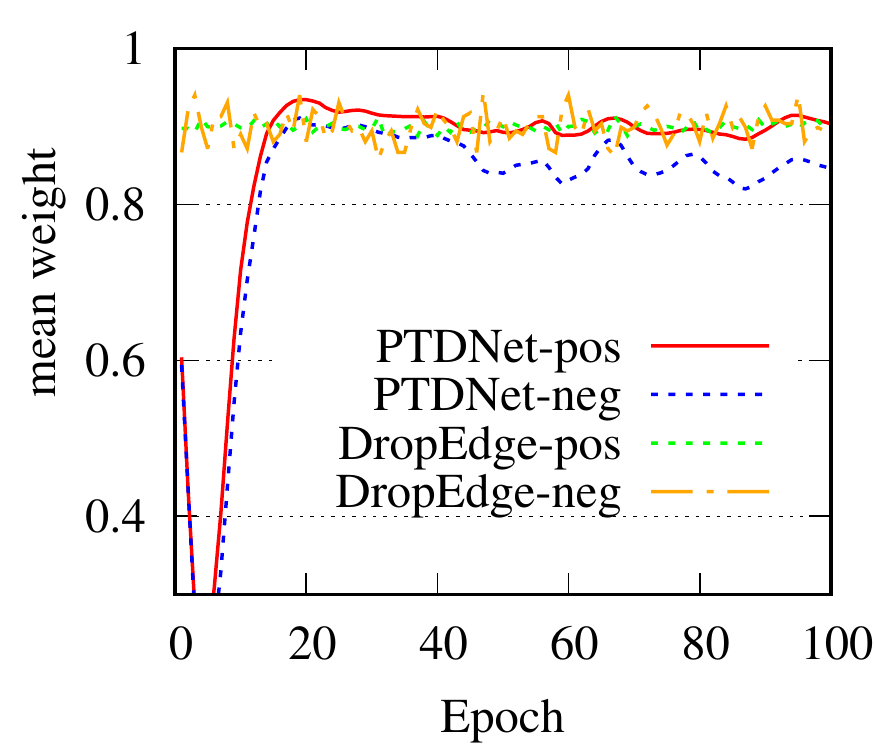}\label{fig:exp3:test}}\hspace{-0.05in}
    \subfigure[{PTDNet accuracy}]{\includegraphics[width=1.4in]{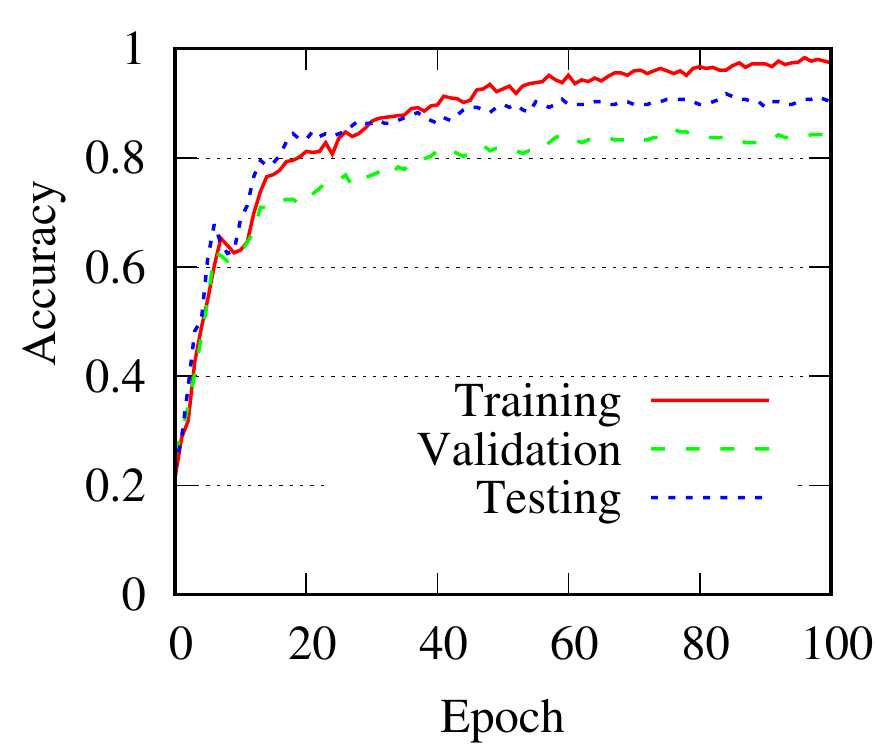}\label{fig:exp3:acc}}\hspace{-0.05in}
     \subfigure[Degree distribution]{\includegraphics[width=1.4in]{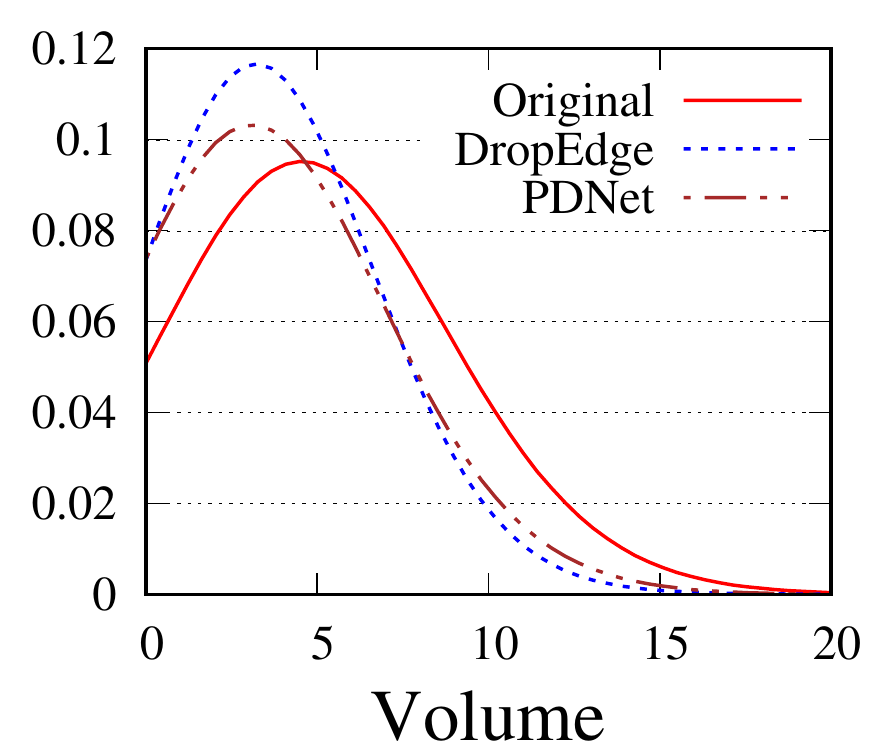}\label{fig:dis:syn}}
         \vspace{-0.3cm}
     \caption{Denoising on a synthetic dataset.}\vspace{-0.3cm}
    \label{fig:exp3}
\end{figure}

We use GCN as the backbone for example. The denoising process of the first PTDNet layer is shown in Fig.~\ref{fig:exp3}. Red lines represent the mean weight, i.e., $z$ in Sec.~\ref{sec:local}, of positive edges (edges connected nodes with the same label), and blue dotted lines are for negative edges. Fig.~\ref{fig:exp3:train},~\ref{fig:exp3:test} count the edges linked with a training/testing node, respectively. We also show the results of DropEdge to see the case of random selection.  These figures demonstrate that DropEdge, an unparameterized method,  cannot actively drop the task-irrelevant edges. While PTDNet can detect negative edges and remove or assign them with lower weights.  The denoising process of PTDNet leads to higher accuracy with more iterations, which is shown in Fig.~\ref{fig:exp3:acc}.  In addition, the consistent performance of PTDNet on testing nodes shows that our parameterized method can still learn to drop negative edges in the testing phase, demonstrating the generalization capacity of PTDNet. Besides, we plot the degree(volume) distribution of the input graph, subgraph sampled by DropEdge and PTDNet in Fig.~\ref{fig:dis:syn}.  We observe that both PTDNet and DropEdge can keep the distribution property of the input graph.

\subsection{Effects of regularizers}
\label{sec:exp4}
\begin{figure}
    \centering
 \subfigure[Effect of $\mathcal{R}_c$]{\includegraphics[width=1.4in]{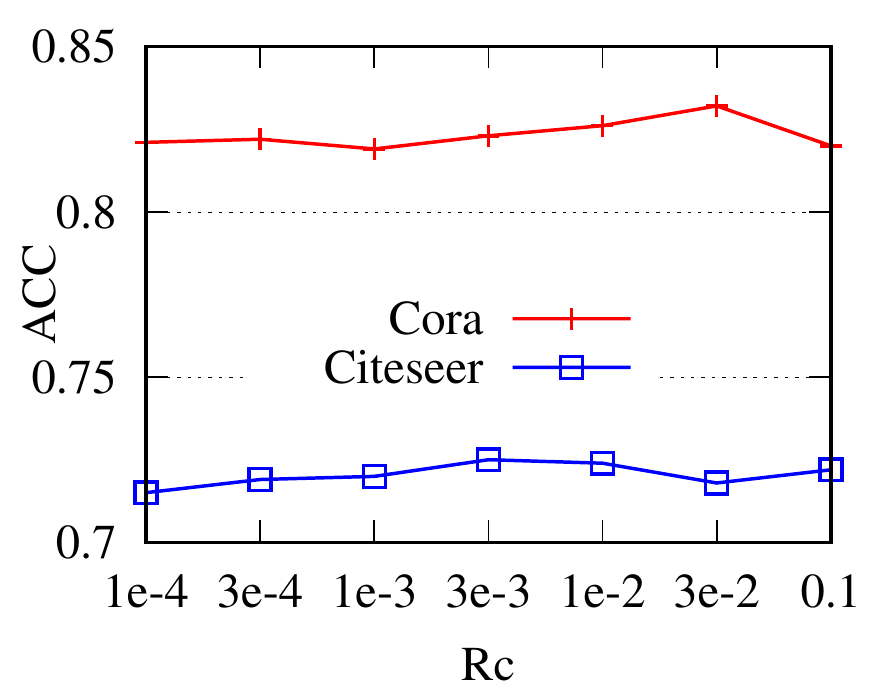}\label{fig:para:paraRc}}\hspace{-0.05in}
 \subfigure[Effect of $\mathcal{R}_{lw}$]{\includegraphics[width=1.4in]{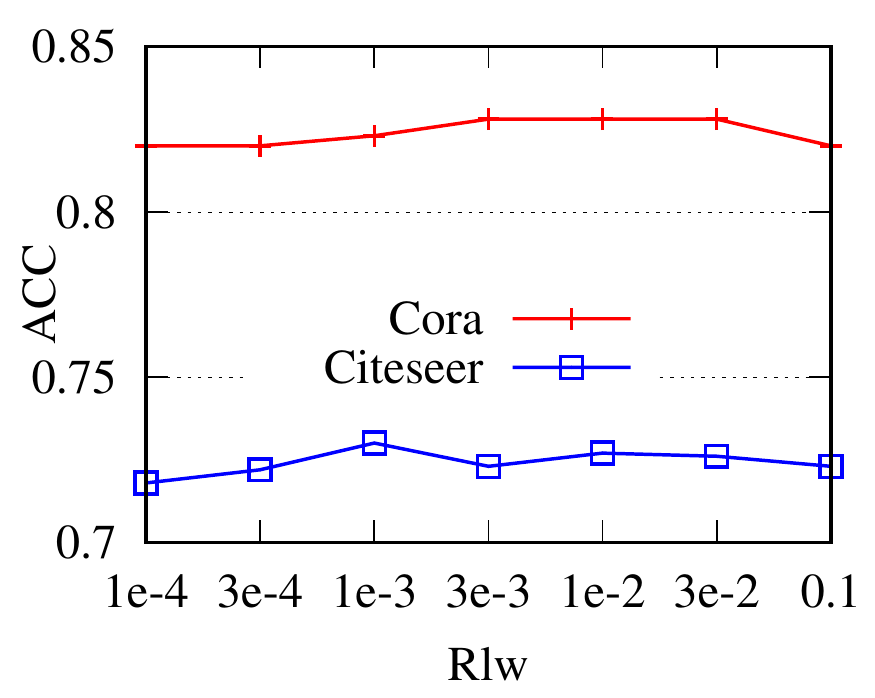}\label{fig:para:paraRlc}}\hspace{-0.05in}
 \vspace{-0.3cm}
       \caption{Effects of $\mathcal{R}_c$ and $\mathcal{R}_{lw}$ on the accuracy.}
    \label{fig:papraReg}
\end{figure}

\begin{figure}
    \centering
 \subfigure[Effect of $\mathcal{R}_c$]{\includegraphics[width=1.4in]{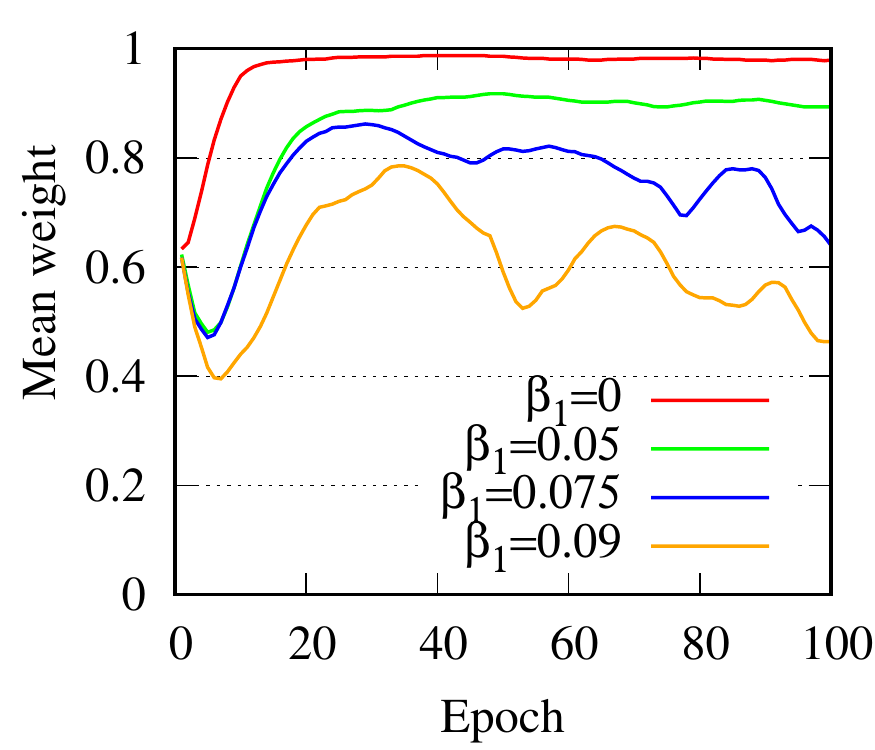}\label{fig:exp4:b1}}\hspace{-0.05in}
 \subfigure[Effect of $\mathcal{R}_{lw}$]{\includegraphics[width=1.4in]{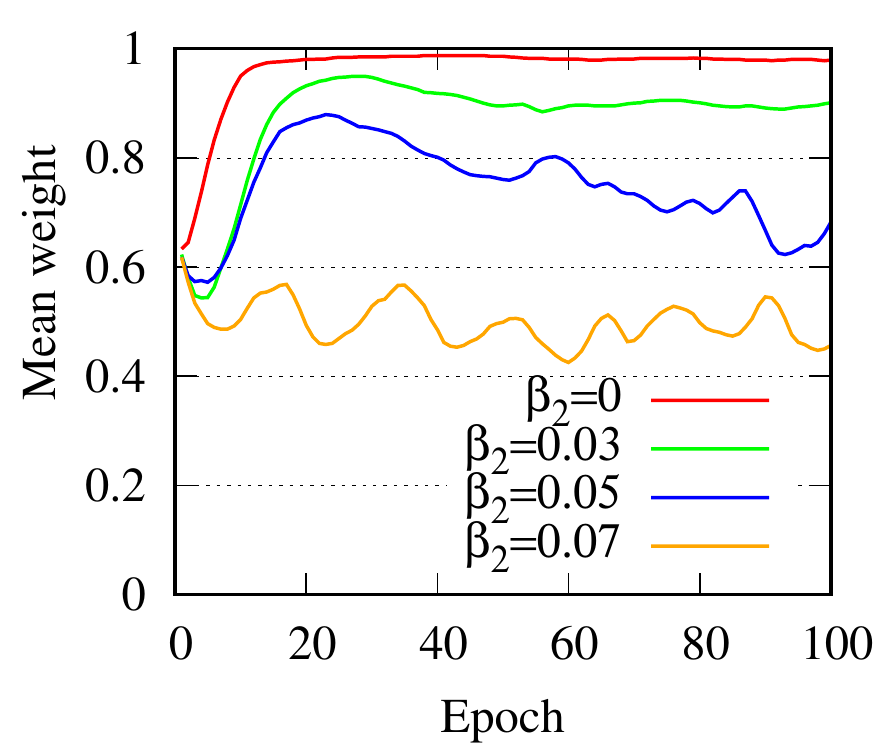}\label{fig:exp4:b2}}\hspace{-0.05in}\vspace{-0.2cm}
       \caption{Effects of $\mathcal{R}_c$ and $\mathcal{R}_{lw}$ on the denoising process.}
    \label{fig:exp4}
   \vspace{-0.3cm}
\end{figure}

In this part, we adopt GCN as the backbone to analyze the effects of regularizers.  We first show the accuracy performance of PTDNet w.r.t coefficients for regularizers in Fig.~\ref{fig:papraReg}. For each choice, we fix that value for the coefficient and use the genetic algorithm to search other hyper-parameters. The best performances are reported here. In general, the performance first increases and then drops as coefficients increase. Since the benchmark datasets are relatively clean. To better show the effects of the proposed two regularizers: $\mathcal{R}_c$ and $\mathcal{R}_{lw}$. We synthesize datasets with controllable properties instead.
We introduce two hyper-parameters $\beta_1$ and $\beta_2$ for regularizers $\mathcal{R}_c$, and $\mathcal{R}_{lw}$, respectively. These two hyper-parameters affect the ratio of edges to be removed. We first dismiss the low-rank constraint by setting $\beta_2=0$. With different choices of $\beta_1$ (0, 0.05, 0.075, 0.9), we show mean weights (i.e., $z$) of edges during iterations in Fig.~\ref{fig:exp4:b1}. The figure shows that with a larger hyper-parameter for $\mathcal{R}_c$, PTDNet achieves a more sparse subgraph. Similar observations can be found in Fig.~\ref{fig:exp4:b2}, where $\beta_1=0$ and only the low-rank constraint is considered.

To demonstrate the effects of including regularizers on the generalization capacity of PTDNet. We synthesize four datasets with various topology properties. The percentages of positive edges in these four datasets range from 0.5 to 0.85. We tune the $\beta_1$ and $\beta_2$ separately by setting the other one to 0. The best options of $\beta_1$ and $\beta_2$ for these four datasets are shown in Table~\ref{tab:exp4}. The table shows that for datasets with poor topological qualities, regularizers should be assigned with higher weights, such that PTDNet can denoise more task-irrelevant edges. On the other hand, for datasets with good structure, i.e., the ratio of positive edges is over 0.85, $\beta_1$ or $\beta_2$ should be with relatively small values to keep more edges. 

\begin{table}[h]
 \vspace{-0.3cm}
    \centering
        \caption{The optimal hyper-parameters for graphs with different qualities.}
    \label{tab:exp4}
    \begin{small}
    \begin{tabular}{c|c|c}
    \hline
        \#pos edge/\#all edges & Best $\beta_1$ & Best $\beta_2$ \\
        \hline
        0.85 &  0.01 & 0.05     \\
        0.7  &  0.04 &  0.07   \\
        0.6  &  0.08 &  0.08    \\
        0.5  &  1.0  &  0.1     \\
        \hline
    \end{tabular}
\end{small}
\end{table}

\normalfont
\begin{table}[h]
\vspace{-0.1cm}
    \centering
    \begin{small}
        \caption{Ratio of cross-community edges in the original graph and subgraphs generated by PTDNets with different settings.}
     \label{tab:tab4}
    \begin{tabular}{c|c|c|c|c}
    \hline
        &original &\begin{tabular}{@{}c@{}} $\beta_1=0.05$ \\ $\beta_2=0 $\end{tabular} & \begin{tabular}{@{}c@{}} $\beta_1=0$ \\ $\beta_2=0.05 $\end{tabular} & \begin{tabular}{@{}c@{}} $\beta_1=0.05$ \\ $\beta_2=0.05 $\end{tabular}\\
        \hline
        $\frac{\text{\#cross comm. eges}}{\text{\#all edges}}$ &0.211 &  0.206 & 0.189  & 0.190  \\
        \hline
    \end{tabular}
    \end{small}
\end{table}
\vspace{-0.3cm}
\normalfont

Low-rank constraint $\mathcal{R}_{lw}$ is introduced to enhance the generalization of PTDNet by setting a constraint on edges connecting nodes from different communities. We use a synthetic dataset to show the influence of $\mathcal{R}_{lw}$. We adopt the spectral clustering to group nodes into 5 communities. As shown in Table~\ref{tab:tab4}, the ratio of cross-community edges is 0.211 in the original graph. We fist adopt PTDNet with $\beta_1=0.05, \beta_2=0$ on the dataset. The ratio in the subgraph generated by the first layer is 0.206. By considering low-rank constraint and setting  $\beta_1=0, \beta_2=0.05$, the ratio drops to 0.189, showing the effectiveness of low-rank constraint on removing cross-community edges.

\subsection{Impacts of approximate factor $K$}
In Sec.~\ref{sec:lowrank}, $K$ is the approximate factor in the low-rank constraint. In this part, we the synthetic dataset used in Sec.~\ref{sec:exp:syn}. We do not utilize the early stop strategies and fix the number of epochs to 300. We range $K$ from 1 to 32 and adopt 2 layer GCN with 256 hidden units as the backbone. The accuracy and running time are shown in Fig.~\ref{fig:expk}.

From the figure, we can see that the accuracy performance increase with the larger $K$, which is consistent with Eq.~\ref{eq:topk} that larger $K$ indicates tighter bound. At the same time, it leads to more running time. Besides, PTDNet can achieve a relatively high performance even when $K$ is small.

\subsection{On over-smoothing}
The over-smoothing problem exists when multiple GNN layers are stacked~\cite{li2018deeper}. DropEdge alleviates this problem by random drop edges during the training phase. Removing certain edges makes the graph more sparse and reduces messages passing along edges~\cite{rong2020dropedge}.  However, As an unparameterized method, DropEdge is not utilized during the testing phase, which limits its power on preventing the over-smoothing problem, especially on very dense graphs. On the other hand, our PTDNet is a parameterized method, which learns to drop in both training and testing phases. 


\begin{figure}[h]
\vspace{-0.3cm}
\begin{minipage}[t]{0.5\linewidth}
    \centering
    \includegraphics[width=1.5in]{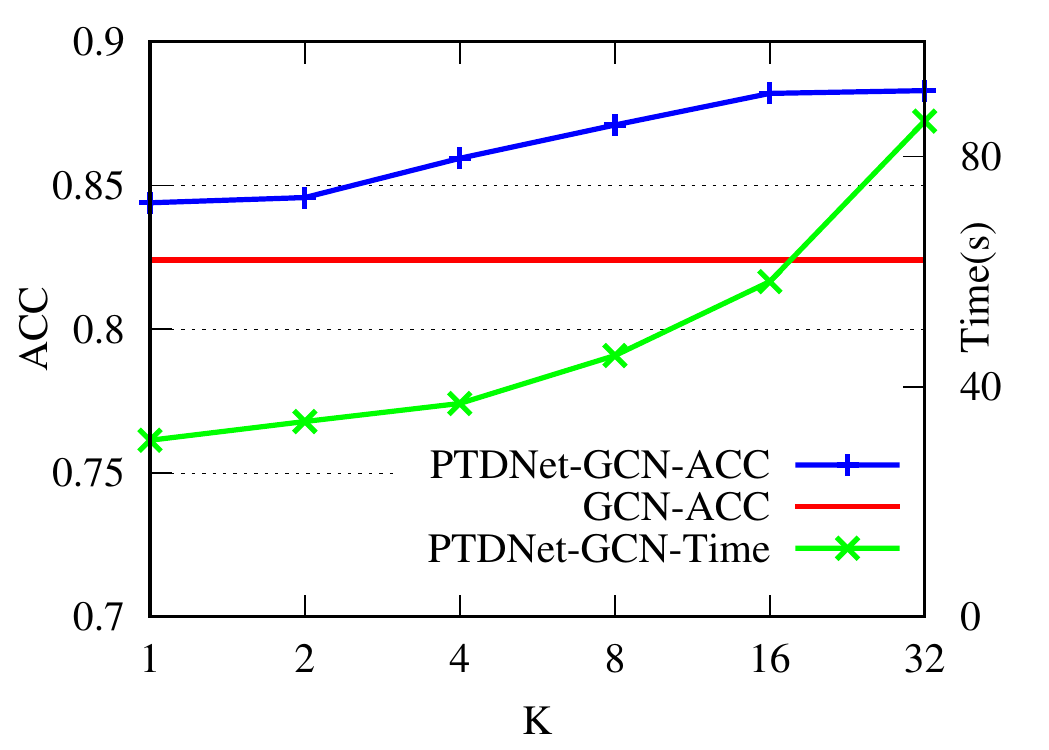}
    \vspace{-0.3cm}
    \caption{Impacts of $K$}
    \label{fig:expk}
\end{minipage}
\hspace{-0.30cm}
\begin{minipage}[t]{0.5\linewidth} 
    \centering
    \includegraphics[width=1.5in]{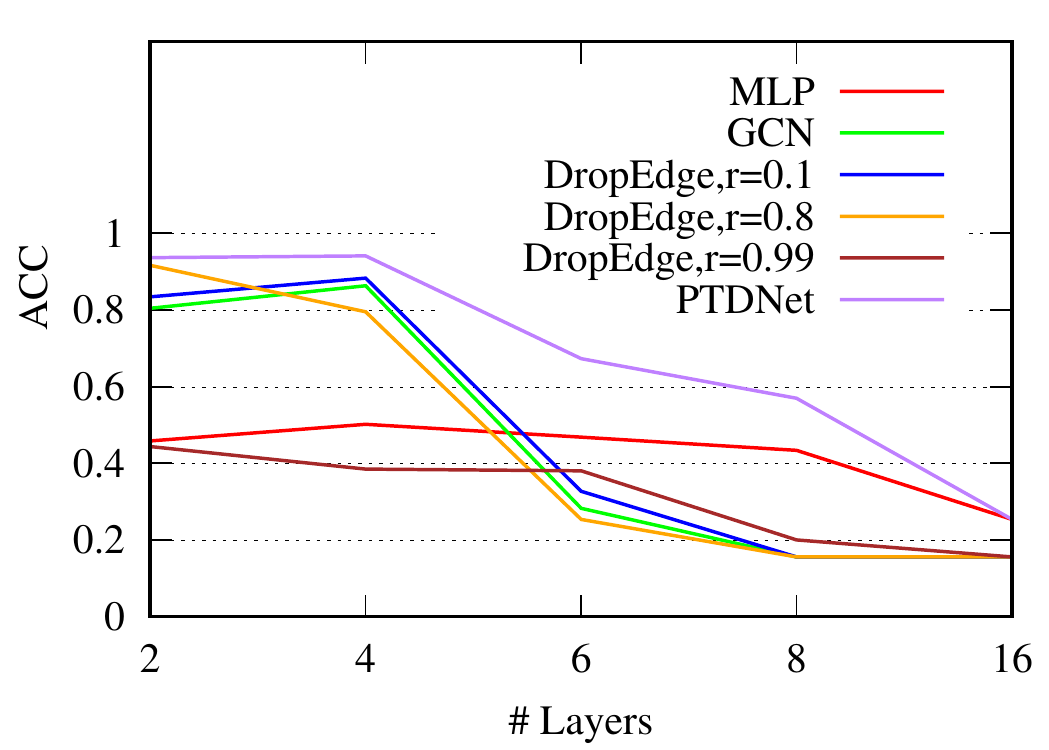}
    \vspace{-0.3cm}
    \caption{On over-smoothing}
    \label{fig:oversmoothing}
\end{minipage}     \vspace{-0.1cm}   
\end{figure}

In this part, we experimentally demonstrate the effectiveness of PTDNet on alleviating the over-smoothing problem in GNN models with a very dense graph.  We adjust the synthetic datasets used in the above section by adding more edges. 
With GCN as the backbone, we compare our PTDNet with DropEdge and the basic backbone. Since stacking multiple GNN also involves the overfitting problem, we include MLP as another baseline, which uses the identity matrix as the adjacency matrix. For DropEdge, we choose three dropedge rates, 0.1, 0.8, and 0.99. We range the number of GCN from 2 to 16 and show the results in Fig~\ref{fig:oversmoothing}.

From the figure, we have the following observations. First, the performances of MLP w.r.t the number of layers show that the overfitting problem appears when 16 layers are stacked. Thus, for models with 8 or fewer layers, we can dismiss the overfitting problem and focus on the over-smoothing merely.  Second, GCN models with 4 layers or more suffer from the over-smoothing problem, which makes the basic GCN model performance even worse than MLP.
Third, DropEdge can only alleviate the over-smoothing problem to some degree due to its limitation on the testing phase. Last but not least, our PTDNet consistently outperforms all baselines. The reason is that our method is parameterized and can learn to drop edges during the training phase. The learned strategies can also be utilized in the testing phase to further reduce the effects of the over-smoothing.

\begin{table}[h]
    \centering
    \begin{small}
        \caption{Performances of Link Prediction}
    \begin{tabular}{c|cc|cc|cc}
    \hline
        \multirow{3}{*}{Method}  & \multicolumn{2}{c}{Cora}  & \multicolumn{2}{c}{Citeseer} & \multicolumn{2}{c}{Pubmed}\\ %
                & AUC   & AP  & AUC   & AP & AUC   & AP\\
        \hline
        GCN(GAE)     &  0.910  & 0.920   & 0.895  &0.899  & \textbf{0.964} &   0.965\\
        DropEdge & 0.881  & 0.903   & 0.862  &0.880  & 0.859 &  0.877\\
        NeuralSparse & 0.901 & 0.917 &0.899 & 0.910 & 0.926 & 0.953 \\
        PTDNet &\textbf{0.916}  &\textbf{0.931}       & \textbf{0.918} &\textbf{0.922}  & 0.963  & \textbf{0.966}\\
    \hline
    \end{tabular}
    \label{tab:linkprediction}
    \end{small}\vspace{-0.3cm}
\end{table}

\subsection{Link prediction}
In this section, we apply our PTDNet to another downstream task, link prediction.
We adopt the Cora, Citeseer, and Pubmed datasets and follow the same experimental settings in GAE, which applies GCN for link prediction~\cite{kipf2016variational}. PPI dataset is not used because it contains multiple graphs and not suitable for link prediction. Specifically, we randomly remove 10\% and  5\% edges for positive samples in the testing and validation sets. The left edges and all node features are used for training.  We include the same number of negative samples as positive edges by randomly sampling unconnected nodes in validation and testing sets.   We adopt area under the ROC curve, denoted by AUC, and average precision, denoted by AP, scores to evaluate their ability to correctly predict the removed edges. GCN is used as the backbone. We compare our PTDNet with the basic GCN, DropEdge and NeuralSparse. Performances are shown in Table~\ref{tab:linkprediction}.

The table shows that our PTDNet can also improve the accuracy performance of link prediction.  The denoising networks in PTDNet are optimized by the downstream task loss and can remove task-irrelevant edges. On the other hand, the performance between DropEdge and the original backbone shows that DropEdge is suitable for the link prediction task.

\section{Conclusion}
\label{sec:conslusion}
In this paper, we propose a Parameterized Topological Denoising Network (PTDNet) to filter out task-specific noisy edges to enhance the robustness and generalization power of GNNs. We directly limit the number of edges in the input graph with parameterized networks. To further improve the generalization capacity, we introduce the nuclear norm regularization to impose the low-rank constraint on the resulting sparsified graphs. PTDNet is compatible with various GNN models, such as GCN, GraphSage, and GAT to improve performance on various tasks. Our experiments demonstrate the effectiveness of PTDNet on both synthetic and benchmark datasets.

\section*{ACKNOWLEDGMENTS}
This project was partially supported by NSF projects IIS-1707548 and CBET-1638320.
\bibliographystyle{ACM-Reference-Format}
\bibliography{ref}
\end{document}